# From WSI-level to Patch-level: Structure Prior Guided Binuclear Cell Fine-grained Detection

Baomin Wang, Geng Hu, Dan Chen, Lihua Hu, Cheng Li, Yu An, Guiping Hu, Guang Jia

*Abstract*—**Accurately and quickly binuclear cell (BC) detection plays a significant role in predicting the risk of leukemia and other malignant tumors. However, manual microscopy counting is time-consuming and lacks objectivity. Moreover, with the limitation of staining quality and diversity of morphology features in BC microscopy whole slide images (WSIs), traditional image processing approaches are helpless. To overcome this challenge, we propose a two-stage detection method inspired by the structure prior of BC based on deep learning, which cascades to implement BCs coarse detection at the WSI-level and fine-grained classification in patch-level. The coarse detection network is a multi-task detection framework based on circular bounding boxes for cells detection, and central key points for nucleus detection. The circle representation reduces the degrees of freedom, mitigates the effect of surrounding impurities compared to usual rectangular boxes and can be rotation invariant in WSI. Detecting key points in the nucleus can assist network perception and be used for unsupervised color layer segmentation in later fine-grained classification. The fine classification network consists of a background region suppression module based on color layer mask supervision and a key region selection module based on a transformer due to its global modeling capability. Additionally, an unsupervised and unpaired cytoplasm generator network is firstly proposed to expand the long-tailed distribution dataset. Finally, experiments are performed on BC multicenter datasets. The proposed BC fine detection method outperforms other benchmarks in almost all the evaluation criteria, providing clarification and support for tasks such as cancer screenings.**

## I. INTRODUCTION

Binuclear cell (BC) detection is an important part of chromosome damage risk assessment in peripheral blood [44]. The occurrence frequency of different types of BCs is closely related to the distinguished diagnosis of an early potential tumor occurrence [1], [2], it is widely used in toxicity screening of carcinogens and early cancer risk evaluation [45].

In the Giemsa-stained microscopy whole slide images (WSIs) of peripheral blood lymphocytes, in addition to the four types of BCs, there are also many other cells as shown in Fig. 1. Detecting these cells has always been challenging due to low-quality staining, the extensive phenomenon of cell nucleus overlapping, slight morphology differences intra- and intertype, etc.

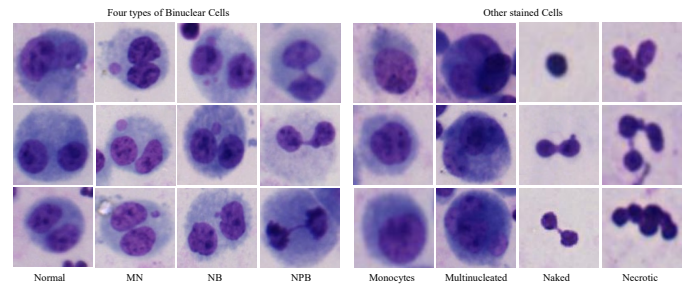

Fig. 1. Examples of different types of cells in peripheral blood lymphocytes microscopy WSIs stained by Giemsa. Normal: normal BCs, MN: micronucleus BCs with outlier small nuclei, NB: nucleus bud BCs with bud-like protuberance, NPB: nucleus bridge BCs with bridging connectors.

At present, BC detection mainly relies on manual microscopy examination, which is time-consuming and lacks objective consistency [3]. Due to the complicated and diverse image features of BCs in the microscopy WSIs, it is difficult to design a precise and efficient general-purpose method based on traditional image processing algorithms for BCs detection [5], [8-11]. Therefore, more robust and generalized methods, such as deep learning-based methods, are needed. There has been some work on BC detection based on deep learning, but these methods have problems such as single-cell type detection [11], or only normal, abnormal coarse BC classification [6], [7], [9], [10], and low recognition accuracy. Most importantly, they do not combine the unique feature information of BCs for fine detection from the WSI-level, and their methods are not rotation invariant [17], [50], and lack certain robustness and generalization, so BCs detection method needs to be further explored.

In this work, to overcome the detection challenges mentioned above, we propose a two-stage strategy guided by the prior BC structure for BC fine detection. We implement the coarse BC

This work was supported in part by the National Natural Science Foundation of China under Grant 82003427, Grant U2004202, and Grant 81872598, in part by the top-notch personnel program of Beihang University under Grant YWF-20-BJ-J-1053. (Corresponding authors: Guiping Hu.)

Baomin Wang, Geng Hu, and Dan Chen are with the School of Biological Science and Medical Engineering, Beihang University, Beijing 100191, China, also with the School of Engineering Medicine and Beijing Advanced Innovation Center for Big Data-Based Precision Medicine, Beihang University, Beijing 100191, China (e-mail: zy2010128@buaa.edu.cn; hugeng007@163.com; chendan2601@163.com).

Lihua Hu is with the Department of Cardiology, Peking University First Hospital, Beijing 100034, China (e-mail: hu_hlh@bjmu.edu.cn).

Cheng Li, Yu An, and Guiping Hu are with the School of Engineering Medicine and Beijing Advanced Innovation Center for Big Data-Based Precision Medicine, Beihang University, Beijing 100191, China. (e-mail: li_cheng@buaa.edu.cn; yuan1989@buaa.edu.cn; hu_hgp@buaa.edu.cn).

Guang Jia is with the Department of Occupational and Environmental Health Sciences, School of Public Health, Peking University, Beijing 100191, China (e-mail: jiaguangjia@bjmu.edu.cn)

detection at the WSI-level first; then, the BC regions are extracted, and finer types of BCs are further classified at the patch-level. The main contributions of this work are summarized as follows:

- Guided by BCs circular shape and dual-nucleus structure, we propose a circle representation-base and multitask detection model for BC coarse detection at the WSI-level.
- Guided by BCs color layer structure, we propose an unsupervised and unpaired cytoplasm generator network for data expansion, a fine-grained classification network consisting of a region suppression and selection module for BC fine classification at the patch-level.

Section 2 introduces the historical work: BC detection and classification, anchor-free detection, and fine-grained classification. Section 3 introduces specific implementation methods. Following is the results section, and finally, the discussion analysis and conclusion are presented.

## II. RELATED WORK

In this section, we briefly review the related work on BC detection and classification and object detection, especially anchor-free frames. Finally, a transformer in fine-grained classification is introduced.

### A. BC Detection and Classification

In recent years, BC detection based on traditional image processing algorithms has emerged. These approaches use algorithms such as improved watershed, seed region growth, and iterative threshold segmentation based on cell size, shape (aspect ratio, relative concave-convex depth, etc.) [8]-[10], color feature rules [4] to extract regions of interest (ROIs). These algorithms are limited by the small size of the outlying small nucleus in MN, poor contrast between nucleus and cytoplasm, micronucleus aggregation (for blood samples after high-dose radiation), stained impurities, and so on. The complicated microscopy WSIs environment makes it difficult to design robust algorithms, and the detection needs to be further speeded up to meet the needs for rapid detection of large samples of omics data.

Recently, deep learning has been applied to detecting BCs. Su et al. [4] proposed a detection model basing YOLO [11] network to detect normal BCs firstly, then removing normal BCs from the microscopy images by cell color layer feature analysis combined with edge detection algorithms. Because traditional image processing algorithms are still used as critical postprocessing for detection, the detection accuracy and speed still need to be improved. Some researchers have performed work on BC classification. Compared with previous work, Xu et al. [6] combined the spatial transformation network with a general convolutional neural network (CNN) to assist identify MN, which can correct the image to fill the entire field of view, create a better environment for the image recognition of abnormal BCs. Besides, due to the lack of abnormal BCs and the high cost of data labeling, this suggests that the idea of transfer learning may be applied to BC classification. Alafif et al. [7] compared various deep CNNs initialed by transfer learning on ImageNet to classify abnormal BC images and normal BC images.

TABLE I
SUMMARY OF UNNORMAL BINUCLEAR CELL DETECTION

| Year | Methods | MN | NB | NPB |
|------|---------|-----|-----|-----|
| 2004 | Aspect ratio [8] | √ | √ | |
| 2010 | Shortest path [11] | | | √ |
| 2010 | Region growing [9] | √ | | |
| 2010 | Watershed [10] | √ | | |
| 2013 | Top-hat [5] | | √ | |
| 2019 | Spatial transform [6] | √ | | |
| 2020 | YOLO, Color layer [4] | √ | √ | |
| 2020 | Transfer learning [7] | √ | | |
| 2022 | Attention in CNN [12] | √ | | |

The above work only focuses on coarse detection [4] or classification [8]-[10], [6], [7]. From Table I, for the detection of abnormal BCs, few works have involved the fine detection of fine-grained types of BCs, and most work has focused on normal BCs or MN detection, which is not comprehensive. In addition, they did not fully combine the unique structural prior information of BCs to implement the identification instead of using a general-purpose neural network that might work well for natural images, so their BC detection and classification do not perform well.

### B. Anchor-free Detection

At present, object detection methods can be divided into handcrafted feature-based and CNN-based methods. The CNN-based detector can automatically extract comprehensive image features and is further divided into anchor-base [13], [31] and anchor-free [14-18] methods.

The main difference between the anchor-free and anchor-based is that it does not need to preset certain prior anchors but uses some key points or a combination of some shape parameters for detection. The multikey points detection method appeared in the early stage. ExtremeNet [14] proposed a network that abandon the detection idea of using key corner points [15], but uses four boundary extreme points for detection, which improves the classification performance. Our current study is motivated by CenterNet [46], which simplifies the object detection task by the center point detection for location, length, and width regression by the way of a heatmap for object scale zooming. CircleNet [17] was further proposed based on CenterNet [46], which detects the glomerulus as a circular object by predicting the center point and radius. Similarly, EllipseNet [18] uses the center point and the long axis and short axis of the ellipse for cardiac ultrasound localization. These two anchor-free methods cleverly combine the prior information of object shape and perform well.

### C. Transformer in Fine-grained Classification

Fine-grained classification for data with slight differences inter classes and obvious differences intra classes has always been challenging. Attention mechanisms solve the fine-grained problem well by guiding the model to focus on the most noteworthy area within objects [26], [29], [36]. Transformers achieve a good performance in language modeling tasks [39], it has been gradually applied in the computer vision field to compensate for the shortcomings of CNN in modeling global semantic relationships [19-20], [27-28]. In vision transformer models, images are divided into small patches. He *et al.* [20] proposed an approach in which if the relationship between patches can be well modeled, the most helpful regions for fine-grained identification could be selected. To guide the model to focus on these key regions, several patches with the largest attention weights after global modeling of the previous multilayer transformer are selected for classification, this idea has been revised to our fine-grained recognition of BCs.

## III. METHODS

Given a WSI of peripheral blood lymphocytes stained by Giemsa, which consists of normal, MN, NPB, and NB four types of BCs (the number of nuclei is two), monocytes (the number of nuclei is one), multinucleated cells (the number of nucleus is greater than two), naked cells (without stained cytoplasm) and other stained impurities. In this study, our objective is to train a model $f(\cdot)$ that can count the number of different types of BCs. Because there are only slight differences in the morphology structure of the four types of BCs, and the number of abnormal BCs is far less than that of the normal, this long-tailed distribution of BCs makes the model difficult to be robust. We propose a two-stage detection framework (Fig. 2) by coarse detection at the WSI-level and fine classification in patch-level. Additionally, we propose a cytoplasm generator network for BCs data expansion as show in Fig. 4.

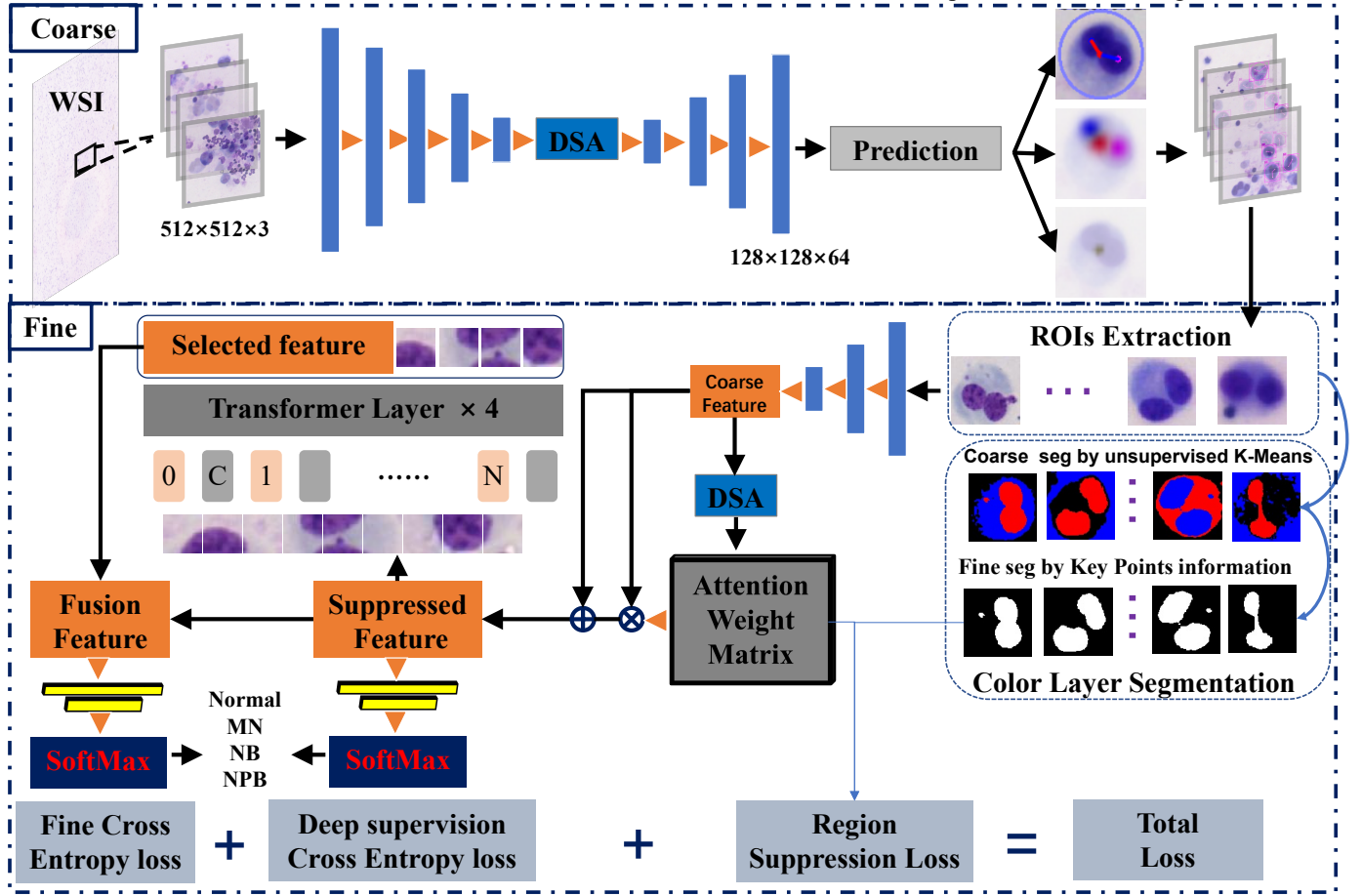

Fig. 2. The overview of our BC detection network. BC image data is extracted from microscopy WSIs stained by Giemsa. The upper part is the coarse BC detection. After the detection, BC coarse ROIs are extracted. Before the second finer classification stage, we use the unsupervised K-Means clustering algorithm for cell color layer segmentation. This mask is used for the region-suppressed module in fine-grained BCs classification, which is made into supervision on the attention feature map. Further, we add a region selection module based on the transformer. Finally, the suppressed and selected features are fused for the final classification.

### A. Circle Key Points Network for BC Coarse Detection

Compare objects to be detected in natural or other medical images, BCs in WSIs have a uniform circular outline structure like the glomerulus, CircleNet [17] is proposed to enable detection of glomerulus with lower degrees of freedom and better rotational consistency by circle representation. Based on CircleNet's [17] direct regression of radius and position coordinates, we further combine the shape structure prior information of BCs by adding the nucleus center key points detection branch to assist the network for better modeling. CKPNet is the proposed BC detection network in the first stage, as described in 'Coarse' part of Fig. 2. Besides, to further expand the receptive field for retention of details, we insert a dilated spatial attention [22] (DSA) module (Fig. 3) between the encoder and decoder.

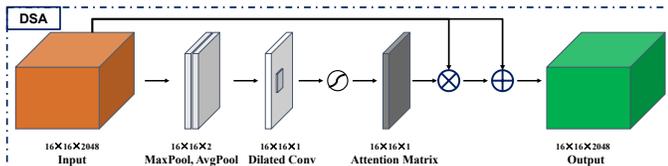

Fig. 3. The inserted dilated spatial attention module.

The overall feature extraction network of CKPNet is similar to U-shaped structure. We use ResNet [21] as the feature encoder. The input RGB image shape is $512 \times 512 \times 3$, and the encoder network consists of 5 stages. In the first stage, the input image goes through a convolutional layer with $stride = 2$, batch normalization, ReLU activation function, and max pooling to a size of $128 \times 128 \times 64$. In the next four parts, the number of residual blocks in each part is $3, 4, 6, 3$, and each residual block contains three convolutional layers. The output feature map size after the encoder is $16 \times 16 \times 2048$, which enters the DSA module. In the DSA module, the input feature map goes through global average pooling $AvgP$ and max pooling $MaxP$ for the features refining, then uses dilated convolution (DC) to refine semantic information such as CBAM [22]. This $3\times3$ dilated convolution makes the spatial attention maps have a bigger receptive field [40], to better adapt to the spatial feature of BCs. The dilated convolution is implemented by a convolution stride of 1, and the padding and dilated ration are both 2 ($DC_{p=2}^{k=3}$), thus the shape of out feature is same as inputs for the next dot multiplied. Finally, the sigmoid ($\sigma$) attention feature map ($F_{attention}$) is dot multiplied and added with the original feature map ($F_{in}$) using a residual connection to obtain the input of the final feature pyramid decoder network ($F_{out}$). The specific implementation is modeled as follows,

$$F_{attention} = \sigma\left(DC_{p=2}^{k=3}([AvgP(F_{in}); MaxP(F_{in})])\right), \quad (1)$$

$$F_{out} = F_{in} + F_{in} \cdot F_{attention}. \quad (2)$$

The decoder network includes cascade 4 blocks of up-sampling and up-convolution. After the decoder, the feature map size is $128 \times 128 \times 64$, and the next is the heatmap prediction. In prediction head, the heatmap is adjusted to the number of object classes along the channel dimension by cascading convolutional layers with kernel size $3 \times 3$ and kernel size $1 \times 1$. For predicting the object center coordinates, the size of output is adjusted to $128 \times 128 \times 2$ for the coordinate offset representation. For predicting the object size, the size of the output is adjusted to $128 \times 128 \times Dof$; in this, $Dof$ represents the degrees of freedom of the object shape [17], [46]. Our BC nucleus key point prediction branch is also parallel here. The specific details of CKPNet are introduced below, and the definitions of some key variables refer to CenterNet [46] and CircleNet [17].

### 1) Circle Representation for BC Detection

Let $I \in R^{W \times H \times 3}$ be the input BC image with width $W$ and height $H$. The output of CKPNet is a heatmap $\hat{Y} \in [0, 1]^{\frac{W}{R} \times \frac{H}{R} \times C}$, where $R$ is the output downsampled stride and $C$ is the number of candidate classes of objects. In the heatmap, $\hat{Y}$ is expected to be 1 at the center of one object and 0 otherwise. For training the object's center point prediction, the ground truth of each key point $p \in \mathcal{R}^2$ is plotted onto a heatmap $Y_{xyc} \in$ $[0, 1]^{\frac{W}{R} \times \frac{H}{R} \times C}$ using a 2D Gaussian kernel [15], [46]:

$$Y_{xyc} = exp\left(-\frac{(x-\tilde{p}_x)^2 + (y-\tilde{p}_y)^2}{2\sigma_p^2}\right) \quad (3)$$

where the $x$ and $y$ are the center point of the ground truth, $\tilde{p}_x$ and $\tilde{p}_y$ are the down-sampled ground truth center point, and $\sigma_p$ is the kernel standard deviation. The training loss is $L_{ob_{hm}}$ penalty-reduced pixel-wise logistic regression with focal loss,

$$L_{ob_{hm}} = \frac{-1}{N} \sum_{xyc} \begin{cases} (1 - \hat{Y}_{xyc})^\alpha \log(\hat{Y}_{xyc}) \ if \ Y_{xyc} = 1 \\ (1 - Y_{xyc})^\beta (\hat{Y}_{xyc})^\alpha \log(1 - \hat{Y}_{xyc}) \ otherwise \end{cases}, \quad (4)$$

where $\alpha$ and $\beta$ are hyper-parameters to the focal loss [47] and $N$ is the number of key points. We set $\alpha = 2$ and $\beta = 4$ in experiments.

Because of the output down-sampled stride, we additionally predict a local offset $\hat{O} \in R^{\frac{W}{R} \times \frac{H}{R} \times 2}$ to refine the discretization error for each key point $p$. The object offset loss $L_{ob_{off}}$ is trained with an L1 loss,

$$L_{ob_{off}} = \frac{1}{N} \sum_p \left| \hat{O}_p - \left(\frac{p_{xy}}{R} - \tilde{p}\right) \right|, \quad (5)$$

where $\frac{p_{xy}}{R}$ denotes the down-sampled location, $\tilde{p}$ is the ground truth of the object center point location, $\frac{p_{xy}}{R} - \tilde{p}$ denotes the ground truth of the offset.

In the predicted heatmap, which location value is greater than or equal to its 8-connected neighbors is the object center. The key point location is given by integer coordinates $(x_i, y_i)$ from $\hat{Y}_{x_iy_ic}$ and $L_{ob_{hm}}$. Then the offset $(\delta\hat{x}_i, \delta\hat{y}_i)$ is obtained from $L_{ob_{off}}$. The object's center point $\hat{p}$ and object's radius $\hat{r}$ are defined as:

$$\hat{p} = (\hat{x}_i + \delta\hat{x}_i, \hat{y}_i + \delta\hat{y}_i), \qquad \hat{r} = \hat{R}_{\hat{x}_i, \hat{y}_i}, \quad (6)$$

where $i$ is one of the objects and $\hat{R} \in \mathcal{R}^{\frac{W}{R} \times \frac{H}{R} \times 1}$ is the prediction of the radius for each object center point $p$. The object radius loss $L_{ob_{radius}}$ is a linear regression with L1 loss,

$$L_{ob_{radius}} = \frac{1}{N} \sum_{i=1}^{N} |\hat{R}_{p_i} - r_i|, \quad (7)$$

where $r_i$ denotes the ground truth of the radius for each circle object $i$. Finally, the overall objective is:

$$L_{object} = L_{ob_{hm}} + \lambda_{ob_{radius}} L_{ob_{radius}} + \lambda_{ob_{off}} L_{ob_{off}}, \quad (8)$$

we fix $\lambda_{ob_{radius}} = 0.1$, $\lambda_{ob_{off}} = 1$.

### 2) Nucleus Key Points Detection

Nucleus key points estimation aims to estimate 2 2D nucleus central locations for each BC. We consider the nucleus central as a $2 \times 2$-dimensional offset of the BC's center point. We directly regress to the joint offset (in pixels) with an L1 loss,

$$L_{kp} = \frac{1}{N} \sum_{k=1}^{N} \left| \hat{O}_{ck} - O_{ck} \right|, \quad (9)$$

where $O_{ck}$ is the ground truth of the offset between the object center point and key point, $\hat{O}_{ck}$ is the prediction of this offset, and N is the number of candidate classes of key points. There are two key points in each BC, and we further estimate these 2 joint heatmaps $\hat{K} \in \mathcal{R}^{\frac{W}{R} \times \frac{H}{R} \times 2}$ with focal loss and local pixel offset to recover the discretization error caused by the output stride analogous to $L_{ob_{hm}}$ and $L_{ob_{off}}$. In general, for the supervision of key points ($L_{keypoints}$), we added the offset loss relative to the object center point ($L_{kp}$), the key point heatmap loss ($L_{kp_{hm}}$), and the coordinate offset loss relative to itself ($L_{kp_{off}}$). The total loss is as follows,

$$L_{keypoints} = L_{kp} + L_{kp_{hm}} + L_{kp_{off}} \quad . \quad (10)$$

The multitask detection loss $L_{total}$ is the sum of $L_{object}$ and $L_{keypoints}$:

$$L_{total} = L_{object} + L_{keypoints} \quad . \quad (11)$$

### B. Unsupervised and Unpaired Cytoplasm Generator Network for Data Generation

Facing the long-tailed distribution of BC data, we propose an unsupervised and unpair cytoplasm generator network (CGNet) to expand the abnormal BCs training set. The data expansion does not directly use general GANs [34] but instead uses an image-style transfer network such as [23-25], [35]. The CGNet framework is as described in Fig. 4. In the stained Giemsa WSIs, there are many cytoplasm-free cells, which we call naked cells as shown in Fig. 1. The size of naked cells is relatively smaller than that of BCs, but the overall nucleus structure is very similar to that of BCs. Therefore, we consider using two generative adversarial networks to construct a cytoplasm generator to realize the style transfer from naked cells (Real A in Fig. 4) to BCs (Real B in Fig. 4), thereby realizing the data expansion of abnormal BCs. We reformulate the generated BCs and the real BCs in an adversarial game, when the discriminator cannot easily distinguish the generated and the real sample, it is a good cytoplasm generator.

In the cytoplasm generator network, the input naked cell image $I \in R^{128 \times 128 \times 3}$ goes through the encoder, style converter and decoder and is transferred to the corresponding BC. In the encoder, three convolutional layers extract higher-level features in turn, and the output feature size after the encoder is $32 \times 32 \times 256$. The style converter consists of 6 residual blocks [21], which combine different features of the image from the encoder output feature and convert these feature vectors of the image from the source domain to the target domain. Each residual block contains two convolutional layers, and the input feature is added to the block output to ensure that the properties of the previous layers can be used for the later layers, so that their output does not deviate too much from the original input. In the style converter, instance normalization is used to prevent instance-specific mean and covariance shift [48]. Let $x \in \mathbb{R}^{B \times C \times W \times H}$ be an input tensor containing a batch of $B$ images, each image with channel $C$, width $W$, and height $H$. The Instance normalization layer can be described as:

$$y_{bijk} = \frac{x_{bijk} - u_{bi}}{\sqrt{\sigma_{bi}^2 + \epsilon}}, u_{bi} = \frac{1}{HW} \sum_{l=1}^{W} \sum_{m=1}^{H} x_{bilm},$$

$$\sigma_{bi}^2 = \frac{1}{HW} \sum_{l=1}^{W} \sum_{m=1}^{H} (x_{bilm} - mu_{bi})^2 \quad , \quad (12)$$

where $x_{bijk}$ denote its $bijk - th$ element, where $k$ and $j$ span spatial dimensions, $i$ is the feature channel, and $b$ is the index of the image in the batch. In the decoder, the conversion of deep features to target domain images is achieved by transposing convolution. Considering that the cycle-consistent topological feature consistency constraints cannot be guaranteed if only a single process from naked nucleus cells to BCs in the network, we design a symmetrical cytoplasm remover network. The overall structure of the style transfer network and some implementation details refer to CycleGAN [25].

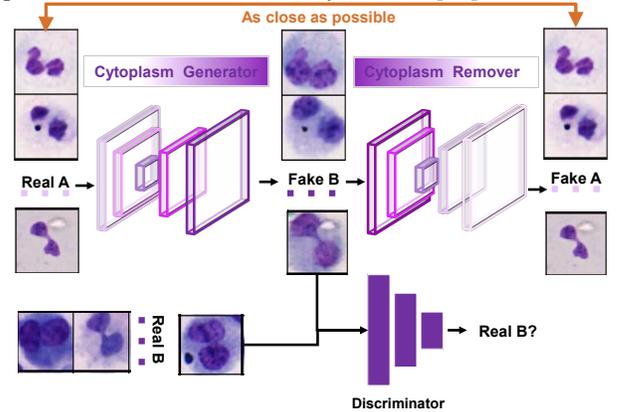

Fig. 4. Overview of our cytoplasm generator and remover network.

### C. Fine-grained Network for BC Classification

BCs have similar structures and shapes, it is difficult to obtain high classification accuracy if the general feature extraction network is used for fine-grained feature extraction. Just like the problem of bone marrow cells classification [26], the 14-category fine-grained classification network proposed based on the attention mechanism better solves the problem of fine-grained cell classification. By an attention mechanism, a neural network is guided to capture fine-grained regions, thus achieve more accurate classification results. In addition, considering the local modeling defects of CNN, BC classification may require transformer global modeling [19].

Our BC fine-grained classification network is a hybrid network of a CNN and a transformer, as described in the 'Fine' part of Fig. 2. CNN is good at extracting shallow local features, and the transformer has better global modeling ability, so the CNN is placed in front of the transformer to accelerate the transformer convergence on small sample data sets [27], [28].

First, it undergoes the preliminary feature extraction of the CNN with a region suppression module to reduce the impact of the staining background and cytoplasm regions on feature learning. Then, the obtained feature map is divided into image patches and then sent to the transformer network with a region selection module for global semantic modeling and fine-grained feature extraction. Finally, we fuse the suppressed feature and the selected feature extracted from Transformer for final classification.

### 1) Region Suppression

The preprocessed BC image $I \in R^{128 \times 128 \times 3}$ enter the CNN branch firstly, then goes through 3 residual convolution layers as described in encoder part of CKPNet, and the out-feature map of the CNN is 1/8 of the size of the original images. The CNN branch consists of three layers of ResNet [21], and every layer has two residual blocks, so it is a lightweight network compared to general ResNet [21] (ResNet-50 has four layers, and each has 3, 4, 6, 3 residual blocks); we call it the R3 network. To guide the subsequent transformer network to pay more attention to meaningful nucleus areas instead of background regions, we designed a region suppression module. The region suppression module is based on the attention matrix, which is the output of DSA. We perform loss supervision on the background region between the outed attention matrix $matrix \in \mathcal{M}^{B \times 1 \times W \times H}$ and the mask $mask \in M^{B \times 1 \times W \times H}$ by unsupervised color layer segmentation. This means that the input matrix and mask both contain a batch of $B$ images, and the width and height are $W$ and $H$, respectively. The mask supervision loss ($Loss_{suppressed}$) is modeled by the L1 loss,

$$Loss_{suppressed} = \frac{1}{N} \sum_{i=1}^{N} |mask_i - matrix_i|, \quad (11)$$

where $N = B \times 1 \times W \times H$, $mask_i$ denotes the output of color layer segmentation, $matrix_i$ represents the output of the attention weight matrix.

Our nucleus region suppression mask is automatically generated by the color layer clustering segmentation method according to the color layer struct feature of BCs. In BC images, there is no staining area, cytoplasm area, or nucleus area, and the first two regions are background. These three regions have different color layers, so the clustering algorithm [49] can easily split them. Next, we choose the color layer of the cell nucleus according to the predicted key points locations generated in the first stage for further finer color layer segmentation, resulting in the binary mask shown in Fig. 2, color layer segmentation. We only use the background mask and the corresponding attention weight matrix for supervision so as not to affect the nucleus region feature extraction by the R3 network and DSA. By the suppression, we make the attention mechanism focus on the cell nucleus area, and the attention weight of other areas is suppressed so that the attention weight after supervision is used as the new attention weight, and the feature map before the suppression module is a dotted product with the new attention matrix weight by residual connection to form the suppressed feature.

### 2) Region Selection

Our region suppression module can be understood as global modeling and then select the most noteworthy regions for feature extraction implemented by Transformer. We added a CNN-based region suppression module before the transformer module mentioned in the previous paragraph to solve the transformer that does not work well on a small data problem. Specifically, this region selection module consists of four transformer layers, and the attention weights of each layer are multiplied to obtain a global attention weight. We select the patches that each head focuses on the most to form the selected feature. The head number of our transformer is eight, so the final feature of the transformer is consists of eight patches. It is based on each attention head of the transformer being considered to focus on features at different levels, and the patch with the highest attention score corresponding to these different heads is the most noteworthy region that needs to be focused on. This part of the transformer-based region selection module refers to TransFG [20].

These selected patches feature is concatenated with the suppressed feature to form the fusion feature for the final classification. The difference between our network with TranFG is that the selected patch features are further concatenated with our previous CNN features for fine-grained feature extraction, similar to some other fine-grained classification networks that have two branches: one for global features and the other for key region features [26], [29]. In addition, it should be noted that deep supervision [30] is used in our classification network and can be calculated as:

$$loss_{cls} = loss_{cnn-cls} + loss_{fusion-cls}. \quad (12)$$

$loss_{cnn-cls}$ and $loss_{fusion-cls}$ are calculated separately after region suppression and feature fusion by a fully connected layer for dimensionality reduction and the softmax joint cross-entropy (CE) loss,

$$CE(s, l) = -\log \frac{\exp (s_l)}{\sum_{i \in J} \exp (s_i)} , \quad (13)$$

where $s$ and $l$ are the confidence scores and the true labels, respectively. $J$ is the set of all categories. The total loss of FBCNet $loss_{total}$ is the sum of $loss_{cls}$ (for classification) and $loss_{suppressed}$ (for region suppression by mask supervision),

$$loss_{total} = loss_{suppressed} + loss_{cls}. \quad (14)$$

## IV. Experimental Details And Results

In this section, we introduce the datasets and data preprocessing methods, and then present the evaluation criteria in the coarse and fine stages. Before introducing the qualitative and quantitative results, we supplement some implementation details.

### A. Dataset

#### 1) Data Source and Preprocessing

Peripheral blood lymphocyte blood sample data were obtained from three medical institutions in China: Chongqing Center for Disease Control (Data Source A), Baotou Medical College (Data Source B), and Henan Institute of Occupational Disease Prevention and Treatment (Data Source C). These blood sample data were processed uniformly by the laboratory of the School of Public Health of Peking University and then made into WSIs. These WSIs are so large that they are difficult to be directly fed into the network for detection or other downstream tasks. Therefore, the sliding window method is used to cut the WSIs into a uniform patch size of $512 \times 512$. In the specific implementation of the sliding window, we add gray bars to the image boundary padding, as shown in the last row of Fig. 6, and when the target is detected, we use the overlapping method to extract the ROIs. These obtained sample data are manually labeled by professional medical experimenters.

#### 2) Data Preparation for Detection

We conduct experiments by multicenter data hybrid and multi-center data cross-validation in two ways. Multicenter hybrid: Hybridize the three data centers' data to make training and test sets according to the ratio. Multicenter cross-validation: select two data centers' data as training sets and other data center as test set. The number distribution of each category in each data center of BCs is shown in Table II. It contains a total of 11,188 BCs consisting of 8,152 normal BCs, 1,536 MNs, 765 NBs, and 735 NPBs. For Multi-center hybrid experiment, we select 9,000 images to form the training set and the validation set, and the remaining 2,188 images are used as the test set according to category quantity weight. For the multicenter cross-validation experiment, we use centers A and B as the training set and center C as the test set.

TABLE II
BINUCLEAR CELLS DATASET

| Data Source | Normal | MN | NB | NPB | Total |
|---|---|---|---|---|---|
| A | 3,215 | 643 | 306 | 416 | 4,580 |
| B | 2,608 | 640 | 313 | 206 | 3,767 |
| C | 2,329 | 253 | 146 | 113 | 2,841 |
| Total | 8,152 | 1,536 | 765 | 735 | 11,188 |

#### 3) Data Preparation for Classification

Our classification data includes data generated by the CGNet. After data expansion, there are a total of 1,421 NBs and 1,103 NPBs. To ensure data balance, we randomly select 1,353 normal BCs as the classification dataset, and the number 1,353 is the mean of the number of MNs, expanded NBs, and NPBs. In all, the classification dataset includes 1,353 normal BCs, 1,536 MNs, 1,421 NBs, and 1,103 NPBs. In particular, the style-transferred data is only used as the training set in classification. For the multicenter hybrid experiment, the number of the test set is 225 in each category, and all the remaining data are used as the training set and validation set (the data division ratio is 4: 1: 1). For the *multicenter cross-validation* experiment, we randomly select 100 images of each category from data center C as the test set and data centers A and B as the training set.

### B. Evaluation Criteria

In BCs coarse detection, we use the MS COCO detection evaluation metrics and F1-score for evaluation. In BCs fine classification, we use the accuracy and confusion matrix for evaluation. The higher accuracy is, the better the overall performance of a classifier is. In two-classification of normal BCs and abnormal BCs, we use the ROC curve, and AUC (Area under the ROC curve), and Sensitivity, and Specificity for criteria. For quantitative evaluation of the BCs generated by the cytoplasm generator network, we use the structural similarity index measure (SSIM) for comparison. SSIM considers the similarity in brightness, contrast, and structure of the two input images $(x, y)$. The closer the calculated SSIM score is to 1, the more similar the two images are. The simplified version of the SSIM calculation formula is as follows [49],

$$SSIM(x, y) = \frac{(2\mu_x\mu_y + C_1)(2\sigma_{xy} + C_2)}{(\mu_x^2 + \mu_y^2 + C_1)(\sigma_x^2 + \sigma_y^2 + C_2)} \quad , \quad (15)$$

where $\mu_x$, $\mu_y$ denote the mean of input images $(x, y)$, $\sigma_x$, $\sigma_y$ denote the standard deviation of input images $(x, y)$, and $C_1$, $C_2$ and $C_3$ are constants that prevent unstable results.

### C. Implementation Details

The coarse detection model in the first stage refers to the PyTorch implementation of CircleNet [17], the cytoplasm generator network refers to CycleGAN [25], and the configuration of each transformer layer in the region-selected module is as follows: the number of heads is eight, the hidden size is 768, and the size of patches is four, referring to ViT [19] and TransFG [20]. All the experiments are performed on NVIDIA Tesla V100 GPUs with 32 GB of memory. We use a random rotation training strategy. For different training tasks, we choose the appropriate iterations epochs until convergence, the batch size is 16 for detection and 32 for classification, the Adam optimizer, and the $1e - 4$ learning rate.

### D. BC Detection Results

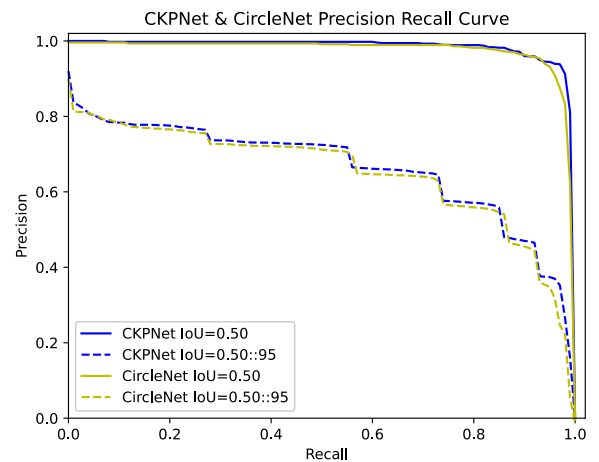

Fig. 5. The precision-recall curve for the detection dataset

In the coarse detection experiments, we compare our proposed CKPNet in multicenter hybrid and cross-validation data split with other networks, performance is as shown in Table III. From the reported results, we can see that in the same backbone (ResNet-50 [21]), our CKPNet is better in AP, AP$_{(50)}$, Recall $_{(50)}$, and F1-score than Faster-RCNN [31] and CenterNet [46]. Compared with CircleNet [17], with ResNet50 as the backbone, our CKPNet has a slight advantage, with an AP value from CircleNet of 0.774 to our 0.778. However, when we replace the backbone with DLA [32] which is a well-designed key point detection network, the key points detection plays a larger role as expected, with an AP value from CircleNet's 0.781 to our 0.798. To some extent, it proves that multitask learning is of some value to the understanding of BC features [42], especially in a well-designed key points backbone.

TABLE III
COARSE DETECTION PERFORMANCE OF BINUCLEAR CELLS

| Dataset | Methods | Backbone | AP | AP $_{(50)}$ | Recall $_{(50)}$ | F1-score |
|---------|---------|----------|-----|------|--------|----------|
| Multicenter hybrid | Faster-RCNN [29] | R50 | 0.621 | 0.840 | 0.833 | 0.836 |
| | CenterNet [46] (Baseline) | R50 | 0.736 | 0.921 | 0.920 | 0.920 |
| | CenterNet+Circle (CircleNet [17]) | R50 | 0.774 | 0.955 | 0.921 | 0.937 |
| | CenterNet+Circle+KeyPoints (Ours) | R50 | 0.778 | 0.959 | 0.920 | 0.939 |
| | CenterNet+Circle (CircleNet [17]) | DLA [32] | 0.781 | 0.956 | 0.926 | 0.940 |
| | CKPNet(Ours) | DLA [32] | 0.798 | 0.959 | 0.933 | 0.946 |
| Multicenter cross-validation | Faster-RCNN [29] | R50 | 0.572 | 0.789 | 0.762 | 0.775 |
| | CenterNet [46] (Baseline) | R50 | 0.681 | 0.854 | 0.870 | 0.861 |
| | CenterNet+Circle (CircleNet [17]) | R50 | 0.720 | 0.915 | 0.875 | 0.894 |
| | CenterNet+Circle+KeyPoints (Ours) | R50 | **0.726** | **0.931** | **0.876** | **0.902** |
| | CKPNet(Ours) | R50-DSA | 0.729 | 0.933 | 0.890 | 0.910 |

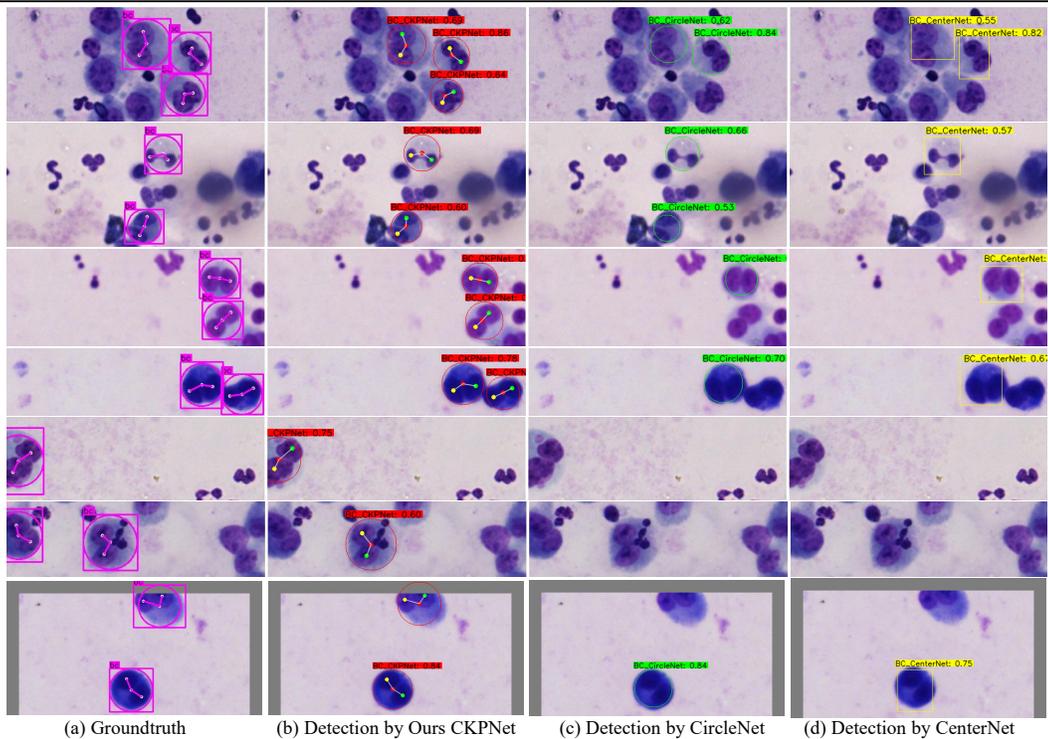

(a) Groundtruth    (b) Detection by Ours CKPNet    (c) Detection by CircleNet    (d) Detection by CenterNet

Fig. 6. Comparison of coarse BC detection results in groundtruth, CKPNet, CircleNet [17], and CenterNet [46]

In CKPNet, we implement the ablation experiments on the dilated spatial attention module by the multicenter cross-validation dataset, it has a further slight improvement, its AP value from 0.726 to 0.729. To further compare the generalization of our method, we compare the two training and testing data combination methods: multicenter cross-validation and multi-center hybrid validation. Compared with the hybrid way, when we only use center A and B data for training and center C for testing in cross-validation experiments, our CKPNet has a certain decline, but compared to other methods such as CenterNet [46] and CircleNet [17], ours is still better. In Fig. 5, we draw the precision-recall curve in different Intersection over Unions (IOUs) to further compare CKPNet and CircleNet [17] in a multicenter hybrid dataset. In the same IOU, CKPNet is almost better than CircleNet [17]. Additionally, we show some BCs detection examples by our CKPNet compared with CenterNet [46] and CircleNet [17] in Fig. 6. Especially in the last row of Fig. 6, when the object is in the gray padding region of the sliding window, our nucleus key points detection can assist in BC detection, but CircleNet [17] and CenterNet [46] cannot.

### E. Cytoplasm Generator Transfer Results

For the cytoplasm generator transfer performance, we randomly select some actual image transfer results, as shown in Fig. 7. The top is the naked cell without cytoplasm, and the bottom is with the cytoplasm. From these images, we find that the topology consistency is maintained well, and the positional relationship of the nucleus to each other is well maintained. Naked cells are covered with a cytoplasm coat after style transfers. The generated images are so good that even professional labeling experimenters can hardly distinguish the reals and the generates.

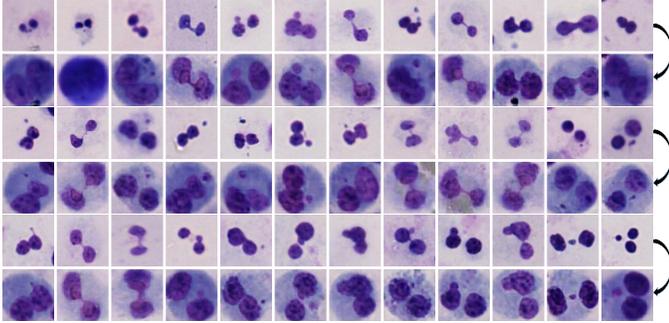

Fig. 7. Examples of the image-style transfer results by our proposed cytoplasm generator network.

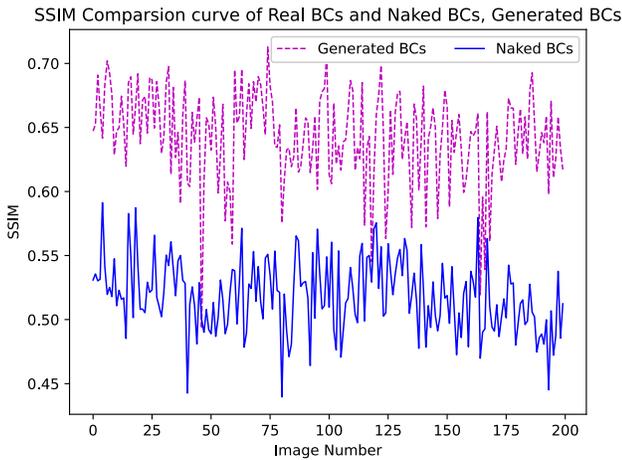

Fig. 8. SSIM comparison curve of generated BCs and naked BCs with real BCs.

For the quantitative evaluation of the generated results, there were no paired real BC target images containing cytoplasm for one-to-one comparison. SSIM can evaluate the structural similarity between two images. We calculate and compare the SSIM values between the original naked cells and the real BC images, and between the generated cell and the real BC images. Specifically, we randomly select 200 images from naked cells and 500 images from real BCs and then calculate the mean SSIM value between each naked cell image and each real BC image. These SSIM value constitute the values on the naked BCs curve in Fig. 8. Correspondingly, the SSIM value of the generated and the real BC images are also calculated. These SSIM value correspond to the values on the generated BCs curve in Fig. 8. From the two curves, it can be found that the generated BC images have a certain degree of increase in the structural similarity to real BC images compared to naked cell images.

### F. Finer BCs Classification Results

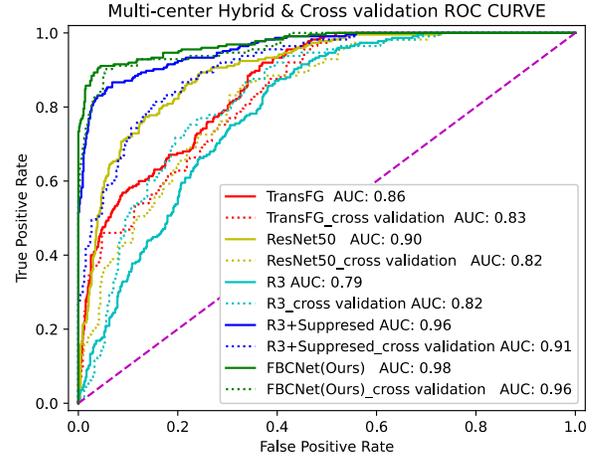

Fig. 9. The classification performance of the two normal and abnormal types of BCs is shown by the ROC curve.

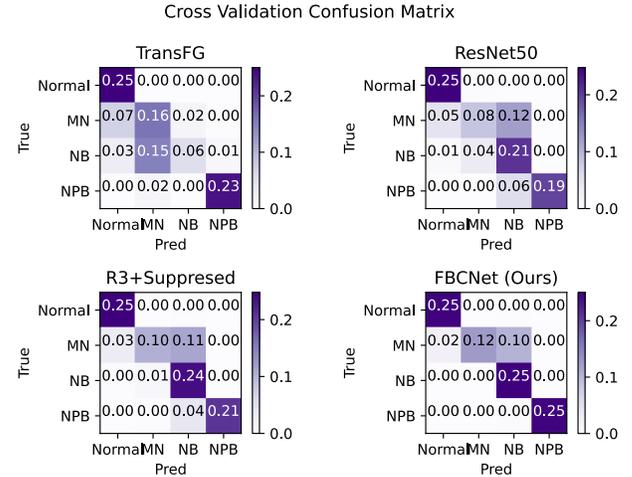

Fig. 10. Four types of BC classification comparison performance are shown by the confusion matrix on the test set. From left to right: TransFG, ResNet50, R3+Suppressed, and R3+Suppressed+Selected

In fine-grained BCs classification, we present the four- and two-classification results with ResNet, TransFG, and FBCNet (Ours) by the performance of AUC, accuracy, sensitivity, specificity, and precision in Table IV. In both kinds of dataset, ResNet and TransFG are both far worse than FBCNet. When we only use the R3 network as the feature extraction network, it has poor performance compared to ResNet-50 [21] and TransFG may because of its' fewer parameters and shallower network depth. When we add the suppressed module and selected module, the performance significantly improves the ablation experiment results. In the multicenter cross-validation dataset, TransFG's four-classification accuracy is 0.7, and our FBCNet has a 0.11 improvement by the region suppressed module and 0.07 improvement by the region selected module. In two-classification experiments, we divide our classification data into normal BCs and abnormal BCs. Thus, the number of abnormal BCs is the sum of MN, NPB, and NB. In the two datasets, each model performs well in the classification of normal and abnormal, and our FBCNet is also superior to the other models. We further draw the ROC curves of FBCNet in two kinds of datasets, as shown in Fig. 9. Its AUC score is 0.96 in the cross-validation dataset, compared to 0.98 in the multicenter hybrid dataset, which has a slight drop.

The comparison of the four-classification confusion matrix in the multicenter cross-validation dataset is shown in Fig. 10. From these four confusion matrices, we find that MN and NB are easy to confuse under the four contrasting models. As shown in Fig. 1, the difference between NB and MN is whether the micronucleus protrudes from the main nucleus or is completely separated from the main nucleus. The morphological differences are small and difficult to distinguish accurately. Our model has higher accuracy in MN prediction than ResNet50 and R3+suppressed but is not as good as TransFG. On the normal, NB and NPB predictions, our model is the best.

TABLE IV
BINUCLEAR CELL FINE CLASSIFICATION PERFORMANCE

| Dataset | Methods | Four-Classification (Normal, MN, NB, NPB) | | | | Two-Classification (Normal & Abnormal) | | | |
|---|---|---|---|---|---|---|---|---|---|
| | | Acc | Pre | Rec | F1-score | AUC | Acc | Sen | Spec |
| Multicenter hybrid | TransFG [18] | 0.72 | 0.75 | 0.73 | 0.70 | 0.86 | 0.83 | 0.77 | 0.99 |
| | ResNet50 [21] | 0.75 | 0.81 | 0.75 | 0.76 | 0.90 | 0.96 | 0.99 | 0.88 |
| | R3 (Baseline) | 0.68 | 0.68 | 0.68 | 0.62 | 0.79 | 0.93 | 0.93 | 0.96 |
| | R3+Suppresed | 0.84 | 0.87 | 0.84 | 0.84 | 0.96 | 0.94 | 0.92 | 1.0 |
| | FBCNet (Ours) | 0.94 | 0.94 | 0.94 | 0.94 | 0.98 | 0.98 | 0.97 | 1.0 |
| Multicenter Cross-validation | TransFG [18] | 0.70 | 0.73 | 0.70 | 0.67 | 0.83 | 0.90 | 0.86 | 0.99 |
| | ResNet50 [21] | 0.72 | 0.76 | 0.73 | 0.72 | 0.82 | 0.94 | 0.92 | 0.99 |
| | R3 (Baseline) | 0.65 | 0.77 | 0.65 | 0.61 | 0.82 | 0.94 | 0.99 | 0.79 |
| | R3+Suppressed | 0.81 | 0.84 | 0.79 | 0.78 | 0.91 | 0.97 | 0.95 | 0.98 |
| | FBCNet (Ours) | 0.88 | 0.90 | 0.87 | 0.86 | 0.96 | 0.98 | 0.96 | 0.99 |

## V. DISCUSSION

In this section, the innovation and characteristics of our proposed coarse-to-fine BCs fine detection method are discussed point by point.

### A. Coarse-to-Fine Detection Strategy

For the detection of BCs with slight morphology differences, the coarse-to-fine detection strategy is proposed. This strategy is drawn from the detection of traffic signs [33], first-level categories such as warnings and prohibitions are detected first, and then is the fine classification of secondary categories such as 'no parking'. Compared with classification, detection requires not only category judgment, but also target positioning, which means that detection models are more difficult to optimize than classification. Therefore, to better improve the final detection accuracy of BCs with the fine-grained feature differences, we choose to first detect coarsely in WSI-level, and then carefully classification in patch-level. In our BCs work, another reason to avoid implementing direct four-class detection is the long-tailed distribution problem. Although we propose a style transfer network for data expansion, it is a cell generation network for simple images of single cells, not for complex images generation of multi cells, because it is easier to generate a single cell patch (our used size is $128 \times 128$) for classification than to generate an image patch (our used size is $512 \times 512$) for detection because a single cell patch has a smaller size and simpler image scene. Although the speed of this two-stage method is relatively slow, and the overall network structure is a bit complicated, compared to previous counting methods like the currently widely used manual microscopy counting, the detection accuracy and speed of this method have both greatly improved.

### B. Circle Contour and Two Cell Nucleus Structure Priors for Detection

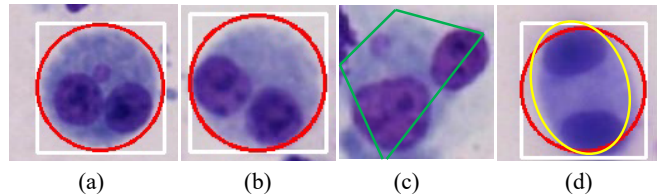

(a)      (b)      (c)      (d)

Fig. 11. Comparison of different bounding boxes.

In the coarse detection stage, combined with cell shape prior information, we propose circular bounding boxes for object representation. Compared with the rectangular bounding boxes which are popular for natural images and have two variables of length and width, circular bounding boxes have only one variable of the radius that needs to be regressed, and they have a lower degree of freedom. Our experiments also demonstrate that using circular bounding boxes has higher detection accuracy in circle BCs detection (the results comparison of CenterNet [46] and CircleNet [17] in TABLE III). Different types of bounding boxes on the BCs visualization effect are shown in Fig. 11. Most shapes of BCs in our WSIs are suborbicular, very few are ellipses, also considering the labeling conveniences and ellipse having two degrees of freedom as a rectangle, we choose circle as the shape of bounding boxes.

The next is the use of two nucleus prior, it was directly found that each BC contained two nucleus, Given that multitask learning can assist features learning [42], we add the detection of these two nucleus together with the whole BC. In data annotation, we set the key points on the left side of the circle center as one type and the key points on the right side as another type. This one-to-one detection supervision ($L_{keypoints}$) assists the CKPNet in correctly recognize the spatial relationship features of dual nucleus and speed up the convergence in the training phase. From Fig. 5 and TABLE III, it can be found that this auxiliary task (key points detection) helps the main task (BCs detection) has higher detection accuracy in the test set.

### C. Cytoplasm Generator Network

The design of CGNet is mainly thanks to CycleGAN [25], an image-style transfer network. Compared with the general GANs [34], [35], CGNet requires fewer data and is more stable in training by the cycle generation structure. In CGNet, the multi-layer residual stacking network is used for cytoplasm generator, cytoplasm remover, and the discriminator network. We initially used a small number of residual blocks, and the transfer results in our BCs transfer test set were not well, but when we used this network for monocytes transfer, we found that it performed well as shown in Fig. 12. Considering that BC images are more complicated than monocyte images, we speculate that it may be due that the perceptual ability of the generator network or the discriminator network is insufficient. Next, more residual network layers were applied for BCs transfer training, and it was found that the results improved, which was also confirmed in our previous speculation. To the best of our knowledge, our cytoplasm generator design has not been mentioned before, and we hope that this idea can be used for abnormal staining quality repair and style transfer of other stained cells.

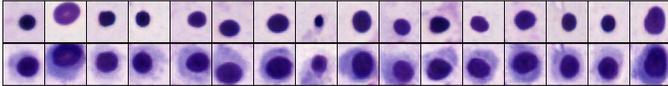

Fig. 12. Cytoplasm generator performance results in monocytes.

Reviewing the design of this style transfer network, we do not have a particularly significant combination of BCs features for further optimization, but only adjusted some parameters and network stacks. Although the results are passable, if we perform some more elaborate design on the implementation details such as targeted network design, our cytoplasm generator may perform better.

### D. Cells Color Layer Priors Guided Fine Classification

Incorporating attention mechanisms into fine-grained classification networks has been proved useful in some state-of-the-arts [29], [36], on this basis, we propose further supervision on the attention map considering that cells have a special color layer structure. For the fine-grained differences in BCs, we speculated that BC classification network requires a powerful global modeling ability, but the general CNNs is lack this ability due to the intrinsic inductive biases, i.e., weight sharing and locality, embedded into its architectural structure [20]. Recently, transformers have been proven to have a better ability to model global relationships than CNNs [19], [28].

However, the transformer is data hungry, and our BCs dataset has less of a training set, so we add three layers of the residual network called R3 network before the transformer network as a shallow feature extractor. Through the residual features extraction, it is found that the transformer network performs well on the small BC dataset, as shown in Fig. 13. From Fig. 13, we can find that when only using Tr (Transformer: ViT base [19]) for BCs detection which is drawn in a green line, it has a slower convergence rate, and its training cross-entropy loss curve is also less stable. When using the R3 network for preliminary feature extraction, such as some state-of-the-art CNN-transformer hybrid networks [27], [28], its convergence rate has a promotion in early epochs, which is shown by the blue line.

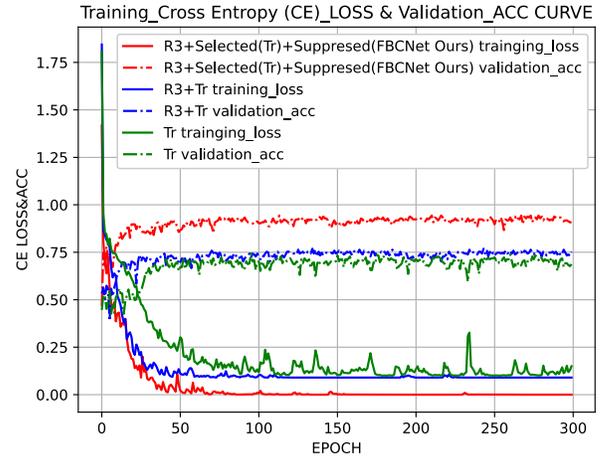

Fig. 13. Our FBCNet training cross-entropy loss and validation accuracy change curve through the epoch. R3: three layers lightweight ResNet, Tr: four layers of the transformer, suppressed: dilated spatial attention module with background supervision module, selected: region-selected module from patch features selected by the transformer.

After shallow feature extraction, we use the spatial attention mechanism mentioned in CKPNet. Considering that BCs have a special shape structure, the entire be classified image has three color layers: no staining layer, cytoplasm layer, and nucleus layer. These color layers can be directly segmented by color layer clustering algorithms, and some examples of segmentation are shown in Fig. 14. We firstly divide the images into three groups of color layers using the K-Means clustering algorithm, which corresponds to the middle row in Fig. 14. Next, considering that our region suppression module only suppresses the background region outside the nucleus, we optimize the color layer segmentation based on the nucleus key point information obtained in the first stage, and we make it contain only the nucleus region and another background area (stained cytoplasm area, unstained area), which corresponds to the bottom row of Fig. 14. Next, the mask is down sampled by the nearest neighbor interpolation algorithm for the attention matrix supervision. Here, we consider that the region outside the nucleus may be of little significance for the overall cell's classification, so the obtained segmentation mask is used to supervise the attention weight map. In particular, it only restrains the region outside the nucleus, not the entire region of BCs, which does not affect the feature map obtained in the nucleus regions. The feature map obtained after the attention guidance module is the input of the transformer network.

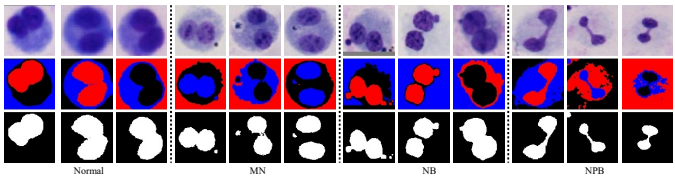

Fig. 14. Examples of color layer segmentation performance.

In the transformer network structure design, we borrow the practice of the fine-grained state-of-the-art transformer network TransFG [20]. After the feature extraction of the multilayer transformer network, we can obtain the focusing area of the multilayer transformer network. We choose the most noteworthy patches concerned by each head to make up the selected feature. We show the abnormal types of BCs in the original images, suppressed feature map in class activation mapping (CAM) [43], and selected patches feature position marked with red dots in Fig. 15. We can find that the suppressed feature and the selected feature attended regions are consistent or complementary. The NPB images of CAM and selected patches may not be consistent as expected, The bridge in NPB is better focused on the region-suppressed CAM, but from the red lines in the selected patches, it is not selected correctly by our region-selected module, the eight selected patches (the number of selected patches depending on the number of heads in each transformer layer mentioned in III. Methods. C) positions are different not like the MN and NB. From Fig. 1, it can be found that NPB has a relatively lightly stained bridging junction, we speculate that this unclear bridge structure has feature loss in deeper layers of the network (transformer), so we implement deep supervision $loss_{cnn-cls}$ by CE loss (13) between the region suppression and selection module.

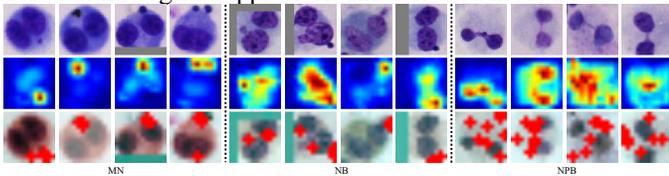

Fig. 15. Original images, the suppressed feature map, and the selected patch positions map.

Finally, the suppressed and selected fused features are used for final classification. In general, the light-weight R3 Network for preliminary feature extraction, suppressed feature extracted by background suppression supervision for reducing background interference on feature extraction, and selected feature extracted by the transformer with its excellent global modeling capabilities for key region feature reinforcement constitute our FBCNet, the final training loss and validation accuracy variation curve with the number of epochs is shown by the red line in Fig. 13. It has a promotion not only in convergence speed but in classification accuracy in the validation set, which is independent of the training set.

## VI. CONCLUSION

In this work, we propose a WSI-level to patch-level detection method, guided by BC special struct priors (circle shape, two nuclei, and color layers), aiming at solving the difficulties faced by BC detection, such as slight morphology differences intra- and inter-types, and long-tailed distribution, thus speeding up the counting of genotoxicity and carcinogenic risk assessment such as leukemia and lymphoma [1-3], [44], [45]. Extensive ablation comparison experiments demonstrated that the circular bounding boxes joint with the nucleus key points coarse detection network, cytoplasm generator network, and fine-grained BCs classification network consisting of background region suppression and key region selection modules can all achieve good performance. Our future work will be to design a more comprehensive and efficient model for the fine detection of more types of stained cells in peripheral blood lymphocyte microscopy WSIs for better auxiliary malignant tumor risk prediction.